\documentclass[conference]{IEEEtran}
\IEEEoverridecommandlockouts
\usepackage{amsmath,amssymb,amsfonts}
\usepackage{algorithmic}
\usepackage{graphicx}
\usepackage{textcomp}
\usepackage{xcolor}
\usepackage{url}

\usepackage{xspace}
\usepackage[numbers,compress,square]{natbib}

\newcommand{\Quant}{\texttt{Quant}\xspace}
\newcommand{\BipolarQuant}{\texttt{BipolarQuant}\xspace}
\newcommand{\Trunc}{\texttt{Trunc}\xspace}
\newcommand{\x}{\texttt{x}\xspace}
\newcommand{\y}{\texttt{y}\xspace}
\newcommand{\scale}{\texttt{scale}\xspace}
\newcommand{\zeropoint}{\texttt{zero\_point}\xspace}
\newcommand{\bitwidth}{\texttt{bit\_width}\xspace}
\newcommand{\signed}{\texttt{signed}\xspace}
\newcommand{\narrow}{\texttt{narrow}\xspace}
\newcommand{\roundingmode}{\texttt{rounding\_mode}\xspace}
\newcommand{\apfixed}{\texttt{ap\_fixed}\xspace}
\newcommand{\apint}{\texttt{ap\_int}\xspace}
\newcommand{\acfixed}{\texttt{ac\_fixed}\xspace}
\newcommand{\acint}{\texttt{ac\_int}\xspace}
\newcommand{\QuantizeLinear}{\texttt{QuantizeLinear}\xspace}
\newcommand{\DequantizeLinear}{\texttt{DequantizeLinear}\xspace}
\newcommand{\Conv}{\texttt{Conv}\xspace}
\newcommand{\MatMul}{\texttt{MatMul}\xspace}
\newcommand{\QLinear}{\texttt{QLinear}\xspace}
\newcommand{\Clip}{\texttt{Clip}\xspace}
\newcommand{\QLinearConv}{\texttt{QLinearConv}\xspace}
\newcommand{\QLinearMatMul}{\texttt{QLinearMatMul}\xspace}
\newcommand{\ConvInteger}{\texttt{ConvInteger}\xspace}
\newcommand{\MatMulInteger}{\texttt{MatMulInteger}\xspace}

\newcommand{\MultiThreshold}{\texttt{MultiThreshold}\xspace}
\newcommand{\quantizedbits}{\texttt{quantized\_bits}\xspace}
\newcommand{\quantizedrelu}{\texttt{quantized\_relu}\xspace}
\newcommand{\QConvTwoD}{\texttt{QConv2D}\xspace}
\newcommand{\QDense}{\texttt{QDense}\xspace}
\newcommand{\QActivation}{\texttt{QActivation}\xspace}
\newcommand{\kernelquantizer}{\texttt{kernel\_quantizer}\xspace}
\newcommand{\biasquantizer}{\texttt{bias\_quantizer}\xspace}

\newcommand{\cnvmod}{\texttt{CNV-w2a2}\xspace}

\DeclareMathOperator*{\clamp}{clamp}
\DeclareMathOperator*{\quantize}{quantize}
\DeclareMathOperator*{\dequantize}{dequantize}
\DeclareMathOperator*{\round}{round}
\DeclareMathOperator*{\hardtanh}{hard tanh}

\begin{document}

\title{QONNX: Representing Arbitrary-Precision Quantized Neural Networks
    \thanks{VL, MT, S-CH, SH, and JD are supported by the US National Science Foundation under Cooperative Agreement OAC-2117997 (A3D3 Institute). 
    JD is also  supported by the US Department of Energy (DOE), Office of Science, Office of High Energy Physics Early Career Research program under Award No. DE-SC0021187 and by the DOE, Office of Advanced Scientific Computing Research under Award No. DE-SC0021396 (FAIR4HEP).
    BH, JMi, JMu, and NT are supported by Fermi Research Alliance, LLC under Contract No. DE-AC02-07CH11359 with the DOE, Office of Science, Office of High Energy Physics and the DOE Early Career Award.
    VL is supported by the European Research Council (ERC) under the European Union's Horizon 2020 research and innovation program (grant agreement n$^o$ 772369), and VL and SS are partially supported under the same program (grant agreement n$^o$ 966696).}
    }

\author{
    \IEEEauthorblockN{Alessandro Pappalardo, Yaman Umuroglu, Michaela Blott}
    \IEEEauthorblockA{\textit{AMD Adaptive and Embedded Computing Group (AECG) Labs} \\
        Dublin, Ireland}
    \and
    \IEEEauthorblockN{Jovan Mitrevski\IEEEauthorrefmark{2}, Ben Hawks, Nhan Tran}
    \IEEEauthorblockA{\textit{Fermi National Accelerator Laboratory} \\
        Batavia, IL, USA}
    \and
    \IEEEauthorblockN{Vladimir Loncar\IEEEauthorrefmark{1}\thanks{\IEEEauthorrefmark{1}Work was partially completed while affiliated with CERN.}}
    \IEEEauthorblockA{\textit{Massachusetts Institute of Technology} \\
        Cambridge, MA, USA}
    \and
    \IEEEauthorblockN{Sioni Summers}
    \IEEEauthorblockA{\textit{European Organization for Nuclear Research (CERN)} \\
        Geneva, Switzerland}
    \and
    \IEEEauthorblockN{Hendrik Borras}
    \IEEEauthorblockA{\textit{Heidelberg University} \\
        Heidelberg, Germany}
    \and
    \IEEEauthorblockN{Jules Muhizi}
    \IEEEauthorblockA{\textit{Harvard University} \\
        Cambridge, MA, USA}
    \and
    \IEEEauthorblockN{Matthew Trahms, Shih-Chieh Hsu, Scott Hauck}
    \IEEEauthorblockA{\textit{University of Washington} \\
        Seattle, WA, USA}
    \and
    \IEEEauthorblockN{Javier Duarte\IEEEauthorrefmark{2}\thanks{\IEEEauthorrefmark{2}Corresponding authors: jduarte@ucsd.edu, jmitrevs@fnal.gov}}
    \IEEEauthorblockA{\textit{University of California San Diego} \\
        La Jolla, CA, USA}
}

\maketitle

\begin{abstract}
    We present extensions to the Open Neural Network Exchange (ONNX) intermediate representation format to represent arbitrary-precision quantized neural networks. 
We first introduce support for low precision quantization in existing ONNX-based quantization formats by leveraging integer clipping, resulting in two new backward-compatible variants: the quantized operator format with clipping and quantize-clip-dequantize (QCDQ) format. 
We then introduce a novel higher-level ONNX format called quantized ONNX (QONNX) that introduces three new operators---\Quant, \BipolarQuant, and \Trunc---in order to represent uniform quantization. 
By keeping the QONNX IR high-level and flexible, we enable targeting a wider variety of platforms. 
We also present utilities for working with QONNX, as well as examples of its usage in the FINN and hls4ml toolchains. 
Finally, we introduce the QONNX model zoo to share low-precision quantized neural networks.
\end{abstract}


\section{Introduction}

Systems for performing neural network (NN) inference often employ reduced numerical precision, potentially with small bit widths, when they are implemented in low-power, low-latency, or high-throughput scenarios.
Such NNs are said to be \emph{quantized} because their weights and activations take on discrete values.
NNs can be trained with floating point values and quantized for implementation after training (post-training quantization or PTQ)~\cite{DBLP:journals/corr/ZhouYGXC17}, or the quantization can be considered during training (quantization-aware training or QAT)~\cite{DBLP:journals/corr/HanMD15,NIPS2015_5647,NIPS2016_6573}.
For example, QKeras~\cite{Coelho:2020zfu,qkeras},  Brevitas~\cite{brevitas}, and HAWQ~\cite{9009512,NEURIPS2020_d77c7035,pmlr-v139-yao21a,hawq} are software frameworks implementing QAT.

Support for representing quantized NNs (QNNs) is widespread across deep learning frameworks and toolchains.
However, the lack of a consensus around how quantization information should be captured at a high level means different toolkits approach the problem differently and interoperability can be a challenge.
The focus of this work is to propose a solution for how to represent QNNs based on the Open Neural Network Exchange (ONNX)~\cite{onnx} format, which is widely used.
We first introduce integer clipping as a way to represent sub-8-bits quantization in ONNX in a backwards-compatible manner that works with existing toolchains, at the cost of some limitations. 
We apply the concept to extend two existing methods to represent quantized neural networks in ONNX, the quantized operator format and the quantize-dequantize (QDQ) format~\cite{onnxruntime}, leading to what we call respectively the quantized operator format with clipping and the quantize-clip-dequantize (QCDQ) format.
We then generalize the concept of QCDQ into a new, more compact and expressive set of operators that we call quantized ONNX (QONNX).
In addition to presenting QCDQ, QONNX, and related software utilities, we show how QAT libraries Brevitas and QKeras can produce QONNX models and field-programmable gate array (FPGA) compiler toolchains FINN~\cite{finn,blott2018finnr} and hls4ml~\cite{Duarte:2018ite,hls4ml} can ingest QONNX models.
\section{Background}
NN quantization is a rich area of research that is broadly concerned with reducing the numerical precision required to represent parameters and activation values within NNs in order to reduce memory, bandwidth and computational requirements at training or inference time.

In this work, we focus on uniform quantization, a generalized fixed-point representation, in the context of NN inference acceleration.
We define the $\quantize$ operator as
\begin{equation}
    \label{eqn:quantize} 
    y = \quantize(x) =  \clamp\left(\round\left(\frac{x}{s} + z\right), y_\mathrm{min}, y_\mathrm{max} \right)\,,
\end{equation}
where $x$ is the floating-point input tensor to quantize, $s$ is the \textit{scale} or \textit{resolution} of the output quantized tensor, also known as \textit{unit of least position} (ULP), $z$ is the \textit{zero-point} or {quantization bias} coefficient. 
The function, $\round$, can be \textit{round-to-even} or \textit{round-to-zero}, and $\clamp$ performs clipping inclusive of the boundaries $y_\mathrm{min}$ and $y_\mathrm{max}$. 
These boundaries are defined as
\begin{align}
    \label{eqn:y_min}
    y_\mathrm{min} & =
    \begin{cases}
        - 2^{n_b - 1} & \text{if signed}\\
        0             & \text{otherwise}
    \end{cases} \\
    \label{eqn:y_max}
    y_\mathrm{max} & =
    \begin{cases}
        2^{n_b - 1} - 1 & \text{if signed}\\
        2^{n_b} - 1     & \text{otherwise}\,,
    \end{cases}
\end{align}
where $n_b \geq 2$ is the bit width of the output quantized tensor, and ``signed'' defines its signedness.
To compute the floating-point representation of a quantized value, we define the $\dequantize$ operator as
\begin{equation}
    \label{eqn:dequantize}
    \dequantize(y) = s  (y - z)\,,
\end{equation}
where $y$ is a quantized tensor, $z$ is its zero-point and $s$ its scale.

Depending on the specifics of how $s$, $z$ and $b$ are defined, different variants of uniform quantization emerge.
The case where $z = 0$ is commonly referred to as \textit{symmetric} quantization, while $z \ne 0$ is referred to as \textit{asymmetric} quantization.
The case where $s$ is a 1D tensor---corresponding to different channels of the input tensor to quantize---is commonly referred to as \textit{channel-wise} quantization, while the case where it is a scalar value is referred to as \emph{layer-wise} or \emph{tensor-wise} quantization.
Traditional fixed-point quantization is obtained by restricting $s$ to a power of two.

Because NN quantization is motivated by increasing efficiency, different kinds of tensors are typically quantized differently to incur the least amount of runtime overhead.
Given a linear layer like a matrix multiplication \MatMul or convolution \Conv, asymmetric quantization is typically adopted for the input because it introduces a bias that can be merged with the layer's bias at compile time, as long as $z$ is restricted to an integer in the range $[y_\mathrm{min}, y_\mathrm{max}]$ to preserve compatibility with zero padding.
Symmetric quantization is typically adopted for weights in order to avoid computing an extra term at runtime in the linear layer.
Channel-wise quantization can be adopted for weights, but not for the input, as it would require intermediate output values of the linear layer to be rescaled before accumulation.
However, this issue is avoided if the layer is depthwise separable.
Biases are typically quantized with $s_\mathrm{bias} = s_\mathrm{weight} s_\mathrm{input}$, so that they can be added directly to the output of the layer without any additional rescaling.
A more in-depth overview of the topic is presented in Ref.~\cite{gholami2021survey}.
\renewcommand{\arraystretch}{1.2}
\begin{table*}[t!]
  \caption{Comparison of different ONNX-based quantized neural network intermediate representations.}
  \label{tab:related}
  \centering
  \begin{tabular}{p{15em}|p{6em}|p{6em}|p{6em}|p{6em}|p{6em}|p{6em}}
                                                         & Arbitrary precision & Rounding variants         & Below 8-bits precision    & Weights-only quantization & Avoid op. duplication & High-precision output \\\hline\hline
    \bf QONNX (this work)                                & \bf\checkmark       & \bf\checkmark             & \bf\checkmark             & \bf\checkmark             & \bf\checkmark         & \bf\checkmark         \\
    \bf QCDQ (this work)                                 & $\times$            & $\times$                  & \bf\checkmark             & \bf\checkmark             & \bf\checkmark         & \bf\checkmark         \\
    \bf Quantized op. with clipping (this work)     & $\times$            & $\times$                  & \bf\checkmark             & $\times$             & $\times$         & $\times$         \\\hline\hline
    QDQ \cite{onnx}                                      & $\times$            & $\times$                  & $\times$             & \bf\checkmark             & \bf\checkmark         & \bf\checkmark         \\
    Integer op.~\cite{onnx}                       & $\times$            & $\times$                  & $\times$                  & $\times$                  & $\times$              & \bf\checkmark         \\
    Quantized op.~\cite{onnx}                       & $\times$            & $\times$                  & $\times$                  & $\times$                  & $\times$              & $\times$              \\
  \end{tabular}
\end{table*}

\section{Related Work}

The ONNX standard intermediate representation (IR)~\cite{onnx} supports three different styles of representing quantized neural networks: the quantized operator format, the integer operator format, and (pseudo)tensor format, also known as QDQ.

In the quantized operator format, an operator like \QLinearConv takes input values for the scale and zero point of input, weights, and output.
Their precise meaning, however, depends on the context. 
The input's scale and zero point are based on the expected quantization of the output of the previous layer, such as \QuantizeLinear, which implements Eq.~\ref{eqn:quantize}, \QLinearConv, or \QLinearMatMul.
In this approach, the input is not quantized by the layer itself.
The weights' scale and zero point are properties of a weight tensor that is already in integer format.
The output's scale and zero point instead reflect a fused requantization operation.
Input, weight, and output tensors are all restricted to 8-bit integer data types.
Quantization of bias in \QLinearConv is implicitly encoded with with a zero point of 0 and a scale equal to the product of the input scale by the weight scale.
The bias's bit width is left as implementation-dependent.

One advantage of the quantized operator representation is that it mirrors a typical fused software backend implementation, so it lessens the requirement for additional compiler graph transformations.
However, there are multiple disadvantages.
At the IR level, there is duplication between floating-point and quantized variants of the same operator (e.g. \Conv and \QLinearConv).
At the graph level, there is duplication in representing input quantization information. 
Scenarios where only weights or activations are quantized, which can be beneficial in some bandwidth-bound contexts, cannot be represented.
Nonlinear activation functions or residual additions computed over high-precision inputs (e.g. 32-bit integer, or int32), which can help in preserving higher quality of results, also cannot be represented, as high-precision outputs are not exposed.

In the integer operator format, an operator like \ConvInteger considers integer input and weight tensors and their respective zero-points, all restricted to int8, while the output is assumed to be int32.
Information around scales is not captured as part of the integer operators, and is left to be represented with separate multiplies.
The advantage of this representation over the \QLinear one is that it does not assume fused output requantization, so higher precision accumulators can be exposed.
However, the same IR duplication issues introduced by the \QLinear operator style persist, as well as the same restriction to scenarios where only weights or activations are quantized.

Finally, the goal of the (pseudo)tensor-oriented approach is to associate quantization information with tensors rather than operators.
To do so, the \QuantizeLinear and \DequantizeLinear are repurposed to represent how a tensor is quantized at a particular point in the ONNX graph.
A float32 tensor is first processed by the \QuantizeLinear operator and then quantized with a certain scale and zero point, yielding an 8-bit integer tensor, and is then immediately dequantized back to float32 with the same scale and zero point with which it was just quantized.
In this way, a QDQed tensor can then be processed by any standard operator with support for float32 inputs.
This style of representation removes duplication in the IR, as it relies on just two operators.
Scenarios where only weights or activations are quantized can be easily represented.
However, going through an explicit quantization step to then immediately remove it is redundant, and the restriction to an 8-bit IR is biased towards byte-aligned backends.

Additionally, as of ONNX opset version 16, a few idiosyncrasies are still present.
Both \QLinearConv and \ConvInteger restrict input quantization to per-tensor scale and zero point parameters, meaning depthwise-separable convolutions with channel-wise input quantization cannot be represented.
\DequantizeLinear supports a 32-bit input, but \QuantizeLinear only supports an 8-bit output, meaning that 32-bit integer values can only be sourced as constants or as outputs of \ConvInteger or \MatMulInteger operators, but not as outputs of a quantization or rerequantization step.
\QuantizeLinear only supports float32 or int32 inputs, meaning that int8 tensors cannot be requantized without going through dequantization first.
Finally, none of the three different approaches summarized can represent tensors quantized to a precision below 8 bits. 

Other quantization representations and hardware deployment strategies are also present in the literature.
In particular, TensorFlow allows quantized representations of tensors similar to, but different from the QDQ approach~\cite{tf_docs_qdq_v2}.
In addition, TensorFlow Lite~\cite{tflite} has an 8-bit quantization scheme appropriate for integer-only hardware implementations~\cite{Jacob_2018_CVPR}.
ONNXRuntime~\cite{onnxruntime} also enables the inference of quantized ONNX models, while TensorRT~\cite{tensorrt} can perform additional optimization of quantized ONNX models.
Ref.~\cite{hanebutte2021prequantized} presents a methodology to separate the quantization process from the hardware-specific model compilation stage via a pre-quantized deep learning model description in standard ONNX format.
Ref.~\cite{jain2020efficient} addresses the disconnect between quantized model representations and efficient hardware implementations by proposing  a new dialect of the Apache TVM deep learning compiler, which can generate efficient code for prequantized models on various hardware platforms.

\section{Representing Low-precision Integers with Clipping}
\label{sec:clipping}
As a first step toward supporting low-precision quantization in ONNX, we propose a scheme using existing operators. 
We achieve this goal by inserting a \Clip after a quantization operation with integer boundaries defined in Eqs.~\ref{eqn:y_min}~and~\ref{eqn:y_max} to implicitly represent a narrower target bit width than the one indicated by the output data type of the quantization operation. 
In the operator-oriented format this means inserting a \Clip directly following \QuantizeLinear, \QLinearConv, or \QLinearMatMul to model a lower precision output quantization.
For lower precision quantized weights and biases, no further steps are necessary as they are already represented as integer tensors, so a precision narrower that their native data type can be imposed by the range of values. 
We call this format the operator-oriented approach with clipping. 

Conversely, for QDQ, we insert a \Clip with integer boundaries between the \QuantizeLinear and \DequantizeLinear operators to model a precision narrower than the native output data type of \QuantizeLinear. 
We term this method quantize-clip-dequantize (QCDQ). 
The limitation of this approach is that because \QuantizeLinear, \QLinearConv, and \QLinearMatMul can currently output 8-bit integers only, it only allows modeling precisions at or below 8 bits. 
Moreover, because \Clip operates over scalar boundaries, more fine-grained quantization schemes, like channel-wise bit width, cannot be modeled.
The advantage, however, is that backward compatibility with existing ONNX-based toolchains is preserved, meaning that quantization below 8 bits can be correctly executed by existing 8-bit backends. 
Future backends with support for below 8-bit acceleration could then exploit these formats by internally fusing the extra clipping and internally generating a narrower output data type.

\section{The QONNX Standard and Software Utilities}
\label{sec:qonnx_standard}
To generically represent a wider class of QNNs, we now introduce an extension to the ONNX standard set of operators, named QONNX, with the goal of providing a high-level representation that can be targeted by training frameworks while minimizing reliance on implementation-specific details.
Table~\ref{tab:related} summarizes the main differences between the formats introduced in this work and existing ONNX-based QNN IRs.
Compared to the formats introduced in the previous section, QONNX aims to sit at a higher level of abstraction. 
Currently, it consists of the following three operators, shown in Table~\ref{table:quant}.
\begin{table*}[t!]
    \caption{The new quantization operators in the QONNX standard format.}
    \label{table:quant}
    \centering
\resizebox{0.77\textwidth}{!}{
    \begin{tabular}{| p{0.9\textwidth} |}
        \hline
        \begin{center}{\Quant: calculate the quantized values of one input tensor and produces one output data tensor.} \end{center}  \\
        Attributes:
        \begin{itemize}
            \item \signed (boolean): defines whether the target quantization interval is signed or not.
            \item \texttt{narrow} (boolean): defines whether the target quantization interval should be narrowed by 1. 
            For example, at 8 bits if \signed is true and \narrow is false, the target is $[-128, 127]$ while if \narrow is true, the target is $[-127, 127]$.
            \item \roundingmode (string): defines how rounding should be computed during quantization.
                  Currently available modes are: \texttt{ROUND}, \texttt{ROUND\_TO\_ZERO}, \texttt{CEIL}, \texttt{FLOOR}, with \texttt{ROUND} implying a round-to-even operation.
        \end{itemize}
        \\
        Inputs:
        \begin{itemize}
            \item \x (float32): input tensor to be quantized.
            \item \scale (float32): positive scale factor with which to compute the quantization.
                  The shape is required to broadcast with \texttt{x}.
            \item \zeropoint (float32): zero-point value with which to compute the quantization.
                  The shape is required to broadcast with \texttt{x}.
            \item \bitwidth (int, float32): the bit width for quantization, which is restricted to be $\ge 2$.
                  The shape is required to broadcast with \texttt{x}.
        \end{itemize}
        \\
        Outputs:
        \begin{itemize}
            \item \y (float32): quantized then dequantized output tensor
        \end{itemize}
        \\\hline
        \begin{center}{\BipolarQuant: calculate the binary quantized values of one input tensor and produces one output data tensor.} \end{center}  \\
        Attributes: None
        \\
        Inputs:
        \begin{itemize}
            \item \x (float32): input tensor to be quantized.
            \item \scale (float32): positive scale factor with which to compute the quantization.
                  The shape is required to broadcast with \texttt{x}.
        \end{itemize}
        \\
        Outputs:
        \begin{itemize}
            \item \y (float32): quantized then dequantized output tensor
        \end{itemize}
        \\\hline
        \begin{center}{\Trunc: truncate the least significant bits (LSBs) of a quantized value, with the input's \scale and \zeropoint preserved.} \end{center}  \\

        Attributes:
        \begin{itemize}
            \item \texttt{rounding\_mode} (string): defines how rounding should be computed during truncation.
                  Currently available modes are: \texttt{ROUND}, \texttt{CEIL}, and \texttt{FLOOR}, with \texttt{FLOOR} being the default.
        \end{itemize}  \\
        Inputs:
        \begin{itemize}
            \item \x (float32): input tensor to quantize.
            \item \scale (float32): positive scale factor with which to compute the quantization.
                  The shape is required to be broadcast with \texttt{x}.
            \item \zeropoint (float32): zero-point value with which to compute the quantization.
                  The shape is required to be broadcast with \texttt{x}.
            \item \texttt{in\_bit\_width} (int, float32): bit-width of the input, which is restricted to be $\ge 2.$
                  The shape is required to broadcast with \x.
            \item \texttt{out\_bit\_width} (int, float32): bit width of the output, which is restricted to be $\ge 2.$
                  The shape is required to broadcast with \x.
        \end{itemize} \\
        Outputs:
        \begin{itemize}
            \item \y (float32): dequantized output tensor.
        \end{itemize} \\\hline
    \end{tabular}}
\end{table*}
The \Quant operator performs uniform affine quantization as defined in Eq.~\ref{eqn:quantize}. 
The \BipolarQuant operator is a variant of the \Quant operator for binary quantization. 
Finally, the \Trunc operator represents truncation of the least significant bits (LSBs) of a quantized value, with the input's \scale and \zeropoint preserved. 
All QONNX operators fuse in a dequantization operation at the output, meaning they take as input a float32 data type and generate as output a float32 data type. 
This is closer to what a training framework would implement (e.g. \texttt{QuantizeAndDequantizeV2} in TensorFlow~\cite{tf_docs_qdq_v2}) and as such it minimizes the burden on framework developers to target it.
Our representation can be seen as a generalized but more compact form of the QDQ and QCDQ formats.

Similar to Ref.~\cite{tf_docs_qdq_v2}, and contrary to QDQ and QCDQ, by fusing the quantize and dequantize operations, we avoid exposing explicitly how the quantized tensor should be represented in integer form, leaving it as an implementation-dependent aspect. 
Lowering to a more implementation-specific format like QDQ, QCDQ, or the quantized operators variants can then be performed at a later stage.

Moreover, we avoid explicitly encoding the concept of tensor-wise versus channel-wise quantization, leaving instead the responsibility to the broadcast semantics among the input tensors \x, \scale, \zeropoint, and \bitwidth.
This simplifies our representation, and at the same time it allows us to represent a wider variety of use cases, including less common ones, e.g. having a tensor-wise scale with a channel-wise bit width.
Similar to QDQ, by defining \scale and \zeropoint as input tensors, we can encode dynamic quantization scenarios where those tensor are computed at runtime on a per-input basis, e.g. having \scale as a function of \x.
We similarly promote the \bitwidth parameter to a tensor to encode scenarios where e.g. the precision of the activations can change on a per-input basis.
Scenarios where \scale, \zeropoint, or \bitwidth do not directly broadcast with \x, e.g. block-wise scaling, can be represented by inserting intermediate tiling and reshaping transformations until broadcasting conditions are met, up to the extreme case where the shape of \scale, \zeropoint, or \bitwidth is equal to shape of \x.

In Eqs.~\ref{eqn:y_min}~and~\ref{eqn:y_max}, we note that the bit width $n_b$ controls the target integer interval that a quantized value is clamped to.
By relaxing $n_b$ to be a float32 value then, we can model quantization to an integer interval that is not aligned to power of two.
This does not change how the quantized values is represented in hardware, e.g. a 7.5-bit value would still require 8 bits.
However, having fine-grained boundaries on the magnitude of the inputs to an operation like a quantized convolution implies fine-grained boundaries on the magnitude of its output.
This, in turn, is relevant to assess whether the operation might overflow given a certain number of output accumulation bits, or if the output can be computed via a fast algorithm like a residue number system~\cite{rns}.
Note that \narrow still applies separately.

Compared to the \Quant operator, the semantics of \Trunc are slightly different.
Similar to the operator-oriented format, the \scale and \zeropoint parameters reflect assumptions on how the input should have been QDQed by a previous layer.
Because no clipping is being modeled, the \narrow and \signed parameters are not necessary.
The \texttt{in\_bit\_width} and \texttt{out\_bit\_width} parameters instead allow one to compute how many LSBs should be truncated.
A typical use case for this operation is the computation of a quantized average pooling variant, where quantized input values are first summed together and then right shifted.

Overall, the QONNX representation is extremely flexible.
In particular, it is straightforward to convert into other quantized NN representations.

\begin{figure}[tbh!]
    \centering
    \includegraphics[trim={20cm 62.5cm 8cm 124cm},clip,scale=0.25]{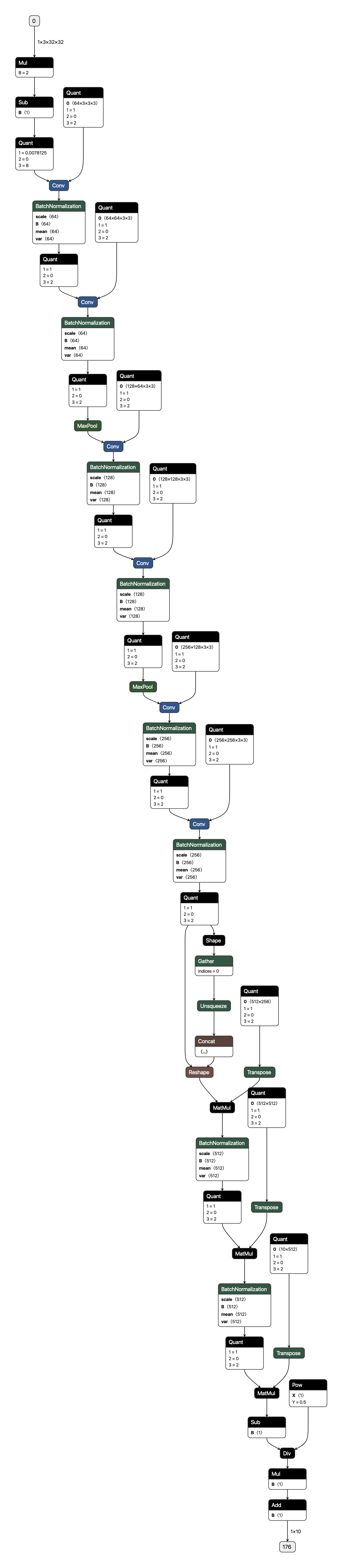}
    \caption{Part of the \cnvmod model taken from the QONNX model zoo, highlighting the transition from convolutional layers to fully connected ones. This model is before any processing.}
    \label{fig:qonnx_raw}
\end{figure}

\begin{figure}[tbh!]
    \centering
    \includegraphics[trim={10.3cm 69.6cm 26cm 156.5cm},clip,scale=0.25]{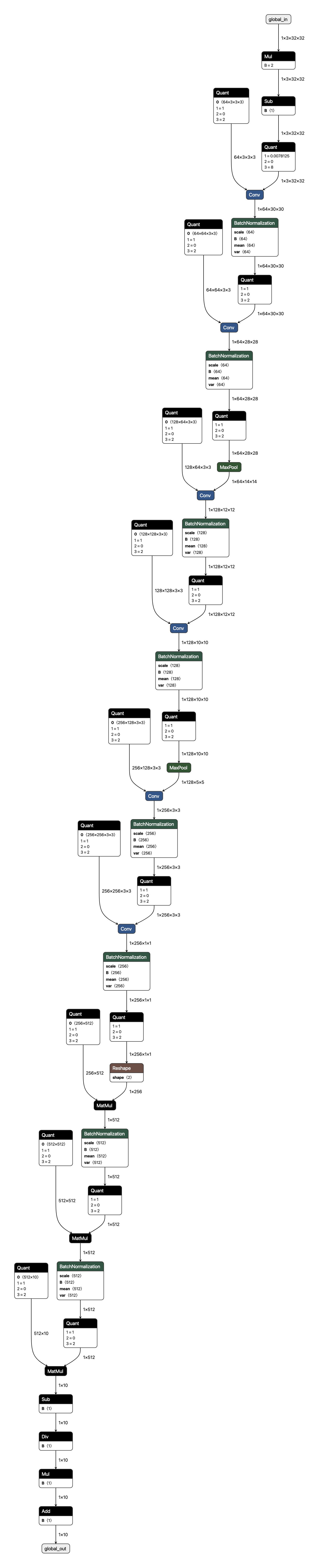}
    \caption{The same part of the \cnvmod model highlighting the transition from convolutional layers to fully connected ones after cleaning. Note that the intermediate tensors now have shape descriptions and the \texttt{Shape}, \texttt{Gather}, \texttt{Unsqueeze}, \texttt{Concat}, and \texttt{Reshape} structure was collapsed into a single \texttt{Reshape} operation.}
    \label{fig:qonnx_clean}
\end{figure}
\begin{figure}[tbh!]
    \centering
    \includegraphics[trim={10.3cm 69.6cm 26cm 156.5cm},clip,scale=0.25]{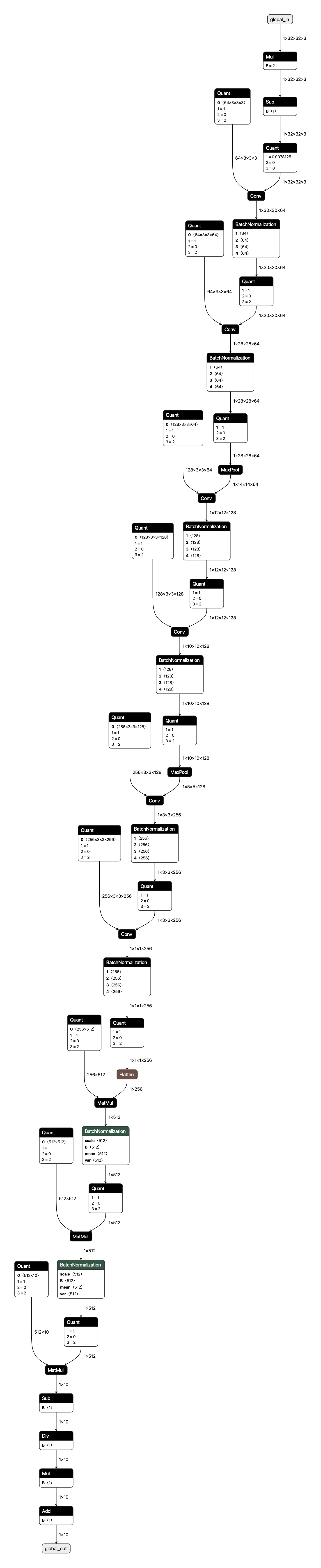}
    \caption{The same part of the \cnvmod model highlighting the transition from convolutional layers to fully connected ones after applying both cleaning and the channels last transformation.}
    \label{fig:qonnx_chan_last}
\end{figure}

To go beyond the pure specification, we also provide utilities for working with QONNX in practice~\cite{qonnx}.
The software utilities currently provide the following features:
\begin{itemize}
    \item Execution of a given QONNX model
    \item Shape inference for intermediate tensors
    \item Basic graph optimizations, such as constant folding
    \item Data layout conversion from a channels-first representation to channels last
    \item ONNX-style documentation for all QONNX nodes\footnote{\url{https://github.com/Xilinx/finn-base/tree/main/docs/QONNX}}
\end{itemize}

All of this functionality is provided as a Python library that can be installed via pip.
Some operations are also available through a command-line interface to make access to the core utilities more convenient to new developers.
The QONNX model execution is based on a node-level execution in Python built with the custom node execution engine used in FINN.
In contrast to other execution engines this one is not meant to provide high performance, but to ensure that model outputs can be verified through execution.

A typical user might start with a network similar to the \cnvmod model found in the QONNX model zoo, discussed in Section~\ref{sec:zoo}.
A part of this network as exported by Brevitas and not yet run through any optimizations is shown in Figure~\ref{fig:qonnx_raw}.
Basic optimizations, such as shape inference and constant folding are generally run before any more involved transformations are applied to the model graph, which the software utilities provide. 
The same \cnvmod model after those cleaning transformations results is shown in Figure~\ref{fig:qonnx_clean}. 
Here, intermediate tensors have shapes attached to them and static nodes were constant folded and have disappeared from the visualization.
A more extensive transformation that a user might run is the data layout conversion.
In standard ONNX, tensors follow the channels-first shape convention of ``batch, channels, height, width'' for 2D data.
However, for both FINN and hls4ml the underlying FPGA implementation expects these tensors to have the channels in the last position, such that the shape becomes ``batch, height, width, channels.''
The QONNX software utilities provide transformations to convert a NN from the channels-first to channels-last format.
Additionally, wrapper nodes exits for shape depended operations, such as convolutions and batch normalization, so that channels last networks can be executed with the FINN execution engine to verify network correctness.
Figure~\ref{fig:qonnx_chan_last} shows once again the example network, now with the data layout conversion applied.
Notably, the 256 channels in the activation tensors have now moved to the last position of the tensor shape.
\section{Integration and Implemented Models}
In this section, we describe the integration of QONNX with existing QAT and FPGA compiler libraries and demonstrate concrete implementations of QONNX models.
\subsection{Conversion of QKeras into QONNX}

The QKeras to QONNX conversion makes use of tf2onnx~\cite{tf2onnx} to perform the actual conversion.
We added handlers to insert \Quant nodes with appropriate attributes based on the quantizers present in the QKeras model.
The conversion is then done in several steps:
\begin{enumerate}    
    \item The model is stripped of QKeras-related attributes (kernel, bias, and activation quantizers) and layers, and replaced with generic Keras layers.
          A map of layers and their quantizers is saved.
    \item The stripped model is converted with tf2onnx.
          The custom layer handlers insert \Quant nodes into the ONNX graph according to the saved map.
    \item We add tensor shape information to the graph initializers and run the cleanup passes previously described.
\end{enumerate}

\begin{figure}
    \centering
    \includegraphics[trim={0 610 0 500},clip,scale=0.25]{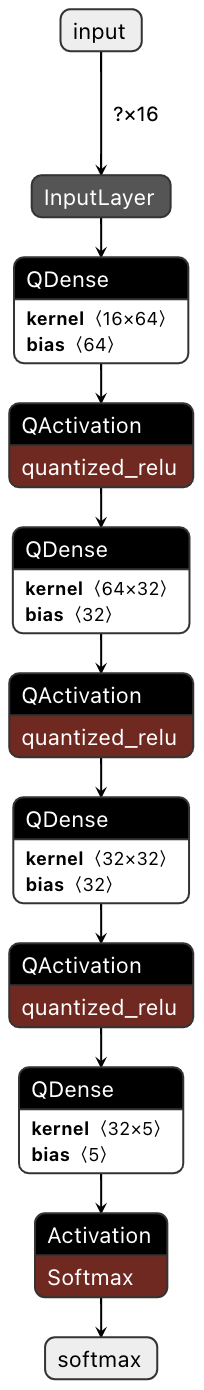}
    \includegraphics[trim={221 1189 169 794},clip,scale=0.25]{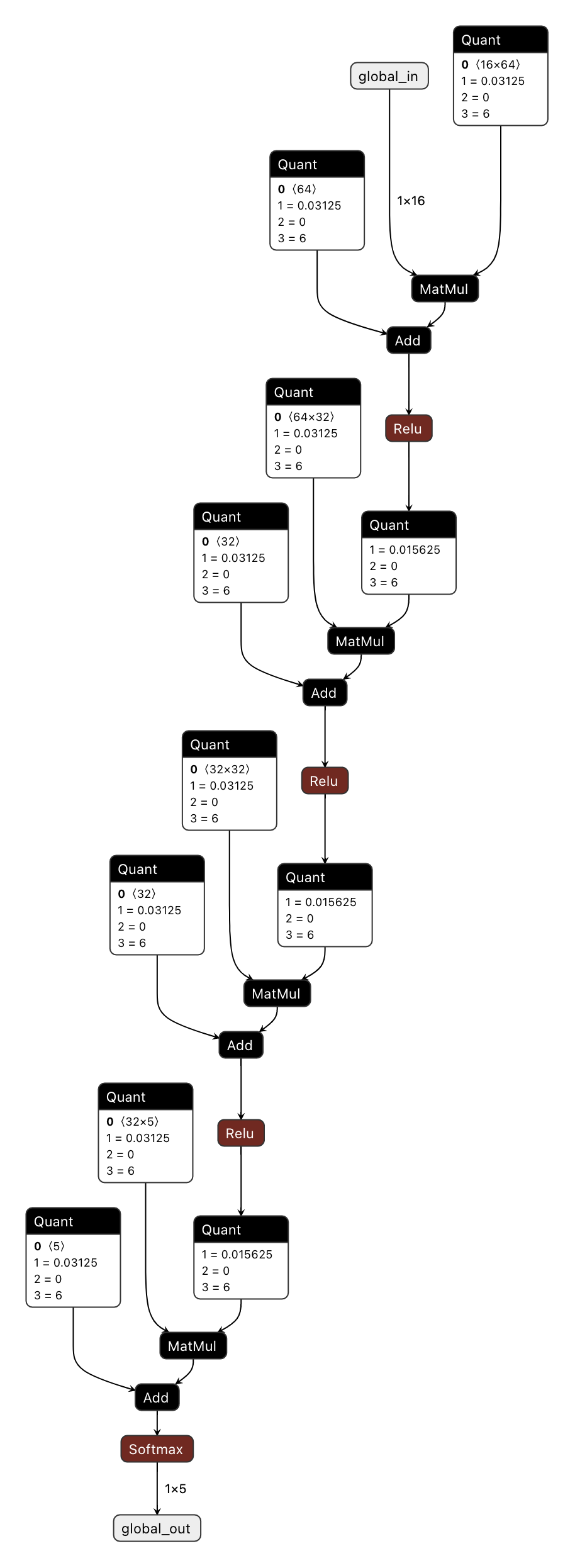}
    \caption{Visualizations of a QKeras model (left) and QONNX representation of the same model (right) after conversion with the QONNX utility.
        The quantizers of the QKeras layers are attributes of the respective layer, and not shown.}
    \label{fig:qonnx_qkeras}
\end{figure}

The \kernelquantizer and \biasquantizer attributes of a QKeras layer, such as \QDense or \QConvTwoD, are transformed to \Quant nodes operating on the respective tensor.
A \QActivation layer is transformed into a standard activation layer followed by a \Quant node.
Currently, the \quantizedbits and \quantizedrelu quantizers of QKeras are supported.
Figure~\ref{fig:qonnx_qkeras} shows an example of the QKeras and QONNX representations of the same model, comprised of a fully-connected layer with quantized weights and biases, followed by a quantized ReLU activation.

\subsection{Brevitas Export of ONNX Variants}

Brevitas generates different ONNX-based formats by extending the existing support in PyTorch for export to ONNX, which is based on tracing the model. 
Because Brevitas implements multiple methods for determining static scales and zero points, at export time their values are first partially evaluated into constants, and traced through with custom autograd functions with symbolic methods matching the desired output ONNX node. 
This allows us to abstract away the underlying algorithms used to determine how the generated quantization operators should be parameterized. 
This approach is adopted for export to QONNX, QCDQ, and the quantized operators format with clipping.

\subsection{hls4ml Ingestion of QONNX}

The hls4ml package~\cite{Duarte:2018ite,hls4ml}, originally developed to deploy QNNs in high energy physics experiments, focuses on easily migrating NNs for implementation in FPGAs, ASICs, and other ``backends.''
The hls4ml framework internally associates a quantization type with every tensor.
These are generally internal representations of \apint and \apfixed~\cite{apfixed} or \acint and \acfixed~\cite{acfixed}.
The precision can be supplied in the configuration, as when doing post-training quantization, or in the model, as when ingesting QONNX or QKeras.
Here we briefly describe how a QONNX input graph is converted to hls4ml's IR.
In all cases, the QONNX graph is first run through the QONNX software utilities for shape inference and constant folding before ingestion.

For simplicity, the \Quant node treatment is split in two categories: that where the \Quant node has unitary scale and zero offset and that where it does not.
In the former case, a \Quant node translates to a quanitzation operation in the hls4ml IR, which takes an input tensor and produces an output tensor with an appropriate quantization type.
If the scale is not unitary or the offset is not zero, a \Quant node translates to three logical operations, first a scaling and shifting operation, then the quantization operation, followed logically by undoing the scaling and shifting (i.e., the dequantization).

To implement the logical operations discussed above more concretely, hls4ml differentiates between quantizations being applied to constants (weights and biases) and quantizations being applied to the data flow.
When the quantization is applied to the data flow, the implementation closely follows what is described above.
When quantizing constants, it is useful to update the constants directly in order to perform the quantization.
More concretely, for the case of a quantization with unitary scale and zero offset being applied to a constant, the constant is simply updated with the quantization applied and the quantization type specified appropriately.
For the case where the scale is not unitary or the offset not zero, the constant is similary updated, but with the scale and offset applied before the quantization; however, in this case a node to dequantize the values is additionally inserted after the constant.

The model graph with the dequantization immediately following the quantization operation can generally not be directly implemented efficiently in hardware, however.
In order to make the graph efficiently synthesizable, the graph needs to be modified.
The dequantization nodes need to be propagated down across linear operators, like matrix multiplications and convolutions so that they can then be done efficiently using quantized values.
The dequantization nodes can be combined with other scalings and shifts, but they may not pass nonlinear activations or quantized nodes, except in certain special cases.

The actual quantization operation is also usually optimized out of the hls4ml IR by incorporating it as the quantization type of the parent's node output tensor.
This is usually safe because care is taken to not cast into the output type prematurely.

One thing we are considering improving is treating power of two scales as a special case.
Currently we do not do anything special, and effectively only use integers values for quantized operations.
However, for power of two scales, we could make use of fractional values in \apfixed and \acfixed representations to effectively propagate the scales through the operations.
Whether the resulting code synthesizes more efficiently versus manually propagating the scales is to be determined.

\subsection{FINN Ingestion of QONNX}

FINN~\cite{finn,blott2018finnr} is a framework to generate streaming neural network hardware for Xilinx FPGAs, optimized for QNNs with few-bit weights and activations.
FINN supports the ingestion of QONNX since version 0.7b, released in November 2021.
To achieve this FINN can automatically detect if a supplied ONNX model contains QONNX nodes and then execute a multistep transformation to convert the QONNX dialect to the internally used FINN-ONNX dialect.

The FINN-ONNX dialect differs significantly from QONNX and is intentionally less flexible as it is optimized for internal use within the compiler toolchain.
There are two notable differences between the ONNX dialects.
The first one being that quantization is expressed as tensor annotations instead of explicit \Quant nodes, thus fundamentally limiting the expressible quantization settings.
The second difference is that instead of using standard ONNX activation functions, FINN implements activations as \MultiThreshold nodes.
These nodes express an arbitrarily quantized activation as a multistep function.
To enable the ingestion of QONNX the basic idea is to convert QONNX to FINN-ONNX.
This is done in four separate steps:
\begin{enumerate}
    \item Using the QONNX software utilities the network is first run through shape inference and constant folding.
    \item Then the weight quantization is applied to the floating point weights and the quantization data type is stored as a tensor annotation.
    \item Following this all \Quant nodes in the activation path are converted to \MultiThreshold nodes.
          Here, FINN currently only supports rectified linear unit, $\hardtanh{}$, and identity activations.
          If an incompatible network architecture is discovered during ingestion an error will be raised.
    \item Finally, special cases such as global average pooling layers are handled in the last step.
\end{enumerate}
With these steps the new ONNX model is then fully converted to the FINN-ONNX dialect and can be used by the rest of the FINN compiler toolchain.
In particular FINN will then start adding new custom nodes to the ONNX graph, which reflect the model implementation in the FINN HLS backend.

\subsection{QONNX Model Zoo}
\label{sec:zoo}
\begin{table*}
    \caption{The models in the QONNX model zoo.}
    \label{table:zoo}
    \centering
    \begin{tabular}{ l l | r r r r r r r r }
        Model name      & Dataset  & Accuracy & Input bits & Weight bits & Act. bits & MACs          & BOPs              & Weights     & Total weight bits \\
        \hline
        MobiletNet-w4a4 & ImageNet & 71.14\%  & 8          & 4           & 4         & 557\,381\,408 & 74\,070\,028\,288 & 4\,208\,224 & 16\,839\,808      \\
        CNV-w1a1        & CIFAR-10 & 84.22\%  & 8          & 1           & 1         & 57\,906\,176  & 107\,672\,576     & 1\,542\,848 & 1\,542\,848       \\
        CNV-w1a2        & CIFAR-10 & 87.80\%  & 8          & 1           & 2         & 57\,906\,176  & 165\,578\,752     & 1\,542\,848 & 1\,542\,848       \\
        CNV-w2a2        & CIFAR-10 & 89.03\%  & 8          & 2           & 2         & 57\,906\,176  & 331\,157\,504     & 1\,542\,848 & 3\,085\,696       \\
        TFC-w1a1        & MNIST    & 93.17\%  & 8          & 1           & 1         & 59\,008       & 59\,008           & 59\,008     & 59\,008           \\
        TFC-w1a2        & MNIST    & 94.79\%  & 8          & 1           & 2         & 59\,008       & 118\,016          & 59\,008     & 59\,008           \\
        TFC-w2a2        & MNIST    & 96.60\%  & 8          & 2           & 2         & 59\,008       & 236\,032          & 59\,008     & 118\,016
    \end{tabular}
\end{table*}

\begin{figure}
    \centering
    \includegraphics[width=0.8\columnwidth]{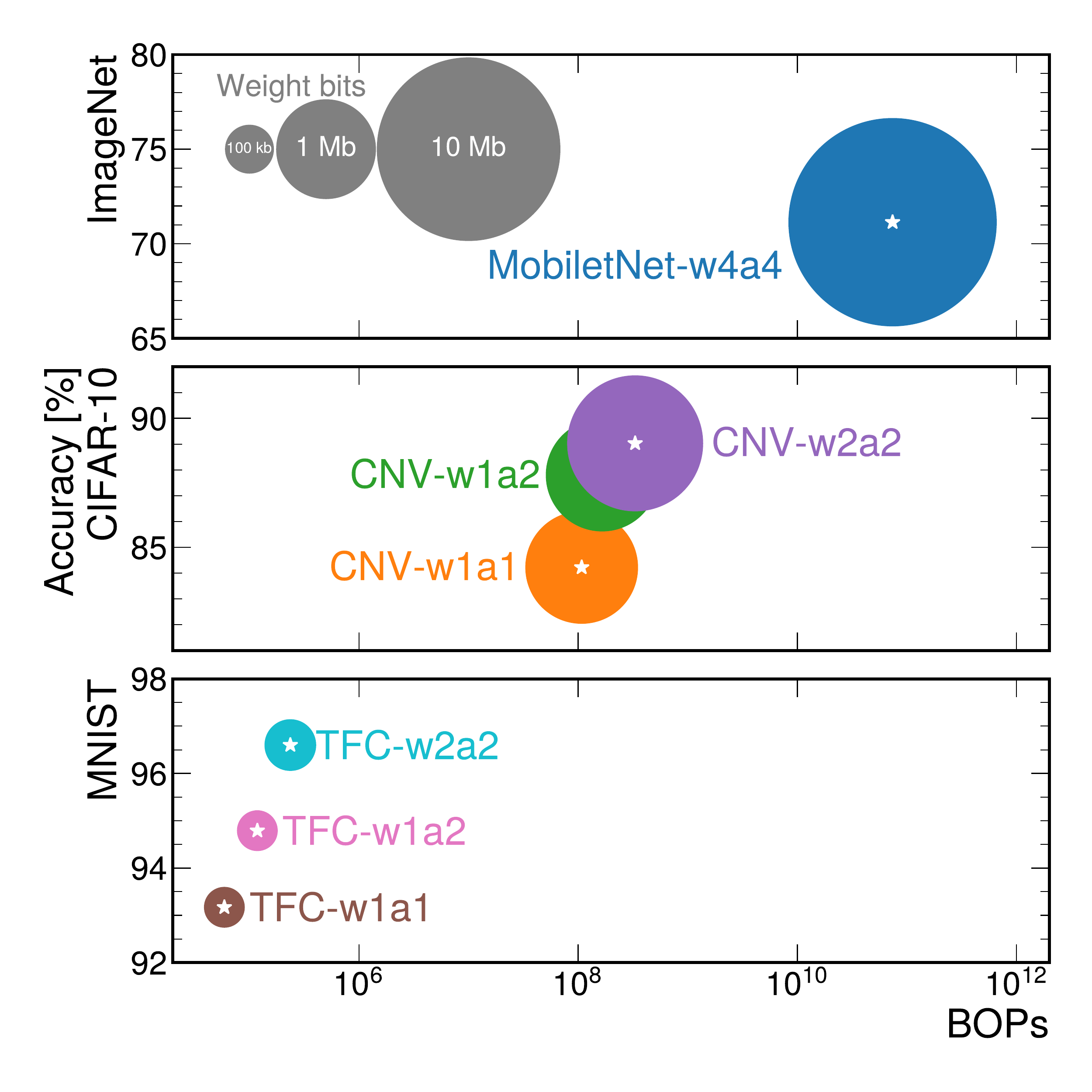}
    \caption{Visual illustration of the QONNX model zoo, depicting accuracy as a function of BOPs, bit operations~\cite{baskin2021uniq}.
        The accuracy metrics are evaluted on either MNIST (bottom), CIFAR-10 (middle), or ImageNet (top), in order of increasing difficulty.
        The size of each marker is a function of the number of total weight bits for each model.}
    \label{fig:qonnx_model_zoo}
\end{figure}

We provide a QONNX model zoo~\cite{modelzoo} with sample QONNX models.
It currently includes:
\begin{itemize}
    \item A MobileNet model based on the MobileNet-V1 architecture~\cite{mobilenet}, trained with Brevitas.
    \item VGG-like small convolutional models (CNV) on CIFAR-10 from \cite{finn}, trained with Brevitas.
    \item Tiny fully-connected models (TFC) on MNIST, with 3 hidden layers of 64 neurons each, trained with Brevitas.
\end{itemize}
Table~\ref{table:zoo} provides more information about these models, which are depicted visually in Figure~\ref{fig:qonnx_model_zoo}.
The model complexity is estimated using metric called bit operations (BOPs)~\cite{baskin2021uniq} that measures the number of bitwise operations in a given network and is defined as:
\begin{equation}
	\mathrm{BOPs} \approx mnk^2( b_\mathrm{a} b_\mathrm{w} + b_\mathrm{a}+b_\mathrm{w} + \log_2 {nk^2})\,,
\label{eq:bops}
\end{equation}
where we consider the BOPs of a single convolutional layer with $b_\mathrm{w}$-bit weights and $b_\mathrm{a}$-bit activations containing $n$ input channels, $m$ output channels, and $k \times k$ filters. 
For computing BOPs for fully connected layers, we set $k=1$ in Eq.~\ref{eq:bops}.  
More information about the models can be found in Ref.~\cite{finn}.
The model zoo is foreseen to grow as more models are converted to QONNX.
\section{Summary and Outlook}

In summary, we propose new representations of quantized neural networks with ONNX. 
We first introduce two low-level formats, quantized operators with clipping and QCDQ, by extending existing formats with the goal of representing below quantization to less than 8 bits in a backward-compatible manner.
We then introduce a novel high-level format called QONNX, which introduces three operators---\Quant, \BipolarQuant, and \Trunc---in order to represent uniform quantization while abstracting away the low-level implementation of quantization, which allows it to target a wider variety of platforms.
It also makes use of broadcast semantics to generalize to tensor-wise and channel-wise quantization.
This paper also presents utilities for working with QONNX, as well as its integration with existing QAT and FPGA compiler libaries.
Finally, we also release the QONNX model zoo.
Given the simplicity and flexibility of this scheme for implementing uniform quantization, a future goal is to present these three operators to the ONNX community for potential inclusion in the ONNX standard itself.

{\footnotesize
    \bibliographystyle{IEEEtran}
    \bibliography{references}
}
\end{document}